# Towards a Technology-Driven Adaptive Decision Support System for Integrated Pavement and Maintenance strategies (TDADSS-IPM): focus on risk assessment framework for climate change adaptation

Shahrzad M. Pour[a,*], Amir M. Masoumi[b], Niels Skov Dujardin[c]

[a]*Technical University of Denmark, Lyngby, Denmark*
[b]*Volunteer researcher and hydraulic engineering expert, Perth, Australia*
[c]*The Danish Road Directorate, Denmark*

**Abstract**

Decision Support Systems for pavement and maintenance strategies have traditionally been designed as silos led to local optimum systems. Moreover, since big data usage didn't exist as result of Industry 4.0 as of today, DSSs were not initially designed adaptive to the sources of uncertainties led to rigid decisions. Motivated by the vulnerability of the road assets to the climate phenomena, this paper takes a visionary step towards introducing a Technology-Driven Adaptive Decision Support System for Integrated Pavement and Maintenance activities called TDADSS-IPM. As part of such DSS, a bottom-up risk assessment model is met via Bayesian Belief Networks (BBN) to realize the actual condition of the Danish roads due to weather condition. Such model fills the gaps in the knowledge domain and develops a platform that can be trained over time, and applied in real-time to the actual event.

## 1. Introduction

With industry 4.0, the request for integrated systems has never been so apparent due to the rapid emergence of big data. Nowadays, transportation infrastructure management systems consist of various System of Systems (SoS). SoS as introduced in Ackoff (1971) consists of suites of interrelated systems, each of which has its own target group in the industry. At the core of this are operation and maintenance planning and activities. In the road sector, traditionally, the pavement and maintenance activities are managed in different physical subunits, leading to isolated systems, even though both systems deal with the same asset; the pavement network. On a higher level, such isolation extended to the development and utilization of Pavement Management Systems (PMS) and Maintenance Management Systems (MMS) as silos. This is an underlying problem stakeholders in this industry face today. A setting that combines these two systems facilitates two opportunities. First, it enables a unified management core that eliminates miscommunication and unaligned measures/decisions. Such unification is achieved by combining the knowledge from Operations Research (OR) on the integrating planning mathematical models, and Operations Management (OM) on the coordination level to achieve a holistic performance and maintenance cost saving.

* Corresponding author. Tel.: +45-5024-3438 ; fax: +0-000-000-0000.
   *E-mail address:* shmp@dtu.dk

The second opportunity comes from the combination of PMS and MMS, which brings adaptive opportunities and real-time assessments. This opportunity is fostered by the emergence of smartness in industry 4.0 as described in Lasi et al. (2014): from reactive/adaptive planning to predictive models and monitoring of KPIs, bridging the gap between advanced Operational Technology (OT) and Information Technology (IT). In the road sector, the climate change and extreme weather events are the most fundamental sources of uncertainty and vulnerability on the asset but also on operation and maintenance services as discussed in Meyer et al. (2013). The importance of design and development of adaptive DSSs to climate change has been the focus of a recent overview paper by Palutikof et al. (2019). In their work, the authors highlight the impact of continuous evaluation of the used DSSs by practitioners, use of adaptive planning, and funding to make such a DSS a reality.

In the literature, at the software level, a few studies introduce the integration of PMS and MMS. Greenstein and Hudson (1998) focus on integration of routine maintenance to PMS back dated 1998. Kachwalla and Hughes (2019) integrates PMS with pavement inspection in real-time manner via GPS and GIS technologies. However no adaptivity to uncertainties exists in the works. Several advancement in PMS in the methods of data collection, data analytics, decision-making, and processing approaches, also applicable for MMS, are reviewed in a comprehensive survey by Peraka and Biligiri (2020). Thanks to all the progresses in various techniques, authors point an immense gap among present research and standpoint application, and highlight necessitating a technology-driven approach as a replacement to the current data-driven approach. To best of our knowledge, there is no research which envisions two holistic aspects of integrating pavement and maintenance planning plus the adaptivity to various source of uncertainties including climate change into one single framework. With a combination of various data sources, we can provide a framework that ensures robustness and resilience immediately and long-term. Accordingly, the contribution of this paper is two folds: i) Introducing the concept of a Technology-Driven Adaptive Decision Support System for integrated pavement and maintenance strategies(TDADSS-IPM) which is adaptive to the sources of uncertainties including climate change. ii) Modeling of the risk assessment module as one component of the proposed framework in order to assess the impact of climate change on road assets. This paper is organized in five sections followed by proposing software architecture in section 2, risk management framework in section 3, and results and conclusion in section 4 and 5, respectively.

## 2. Visionary TDADSS-IPM: Proposed Software Architecture

Figure 1 represents the proposed software architecture for the TDADSS-IPM. The architecture is composed of six components of data integration, data gateway, data ingestion, integrated analytics, prediction and planning, data and information exchange gateway, and applications shown in grey color boxes. Data sources are shown in red. Green boxes refer to single modules each responsible for a certain concept or one business capability covering the activities in relation to PMS, MMS, and adaptivity aspect in an integrated manner. Accordingly, *Notifier* module is the hub/gateway between data sources and the rest of the architecture. Once there is a new flow of data, it will be triggered, and consequently it will inform *Dispatcher* module, where it decides how the new flow should be ingested to the relevant modules (for example Data Persisting or Data Processing, or both of them). *Dispatcher* either ingests the data via batch processing or online/real-time processing depending on the certain business logic per data flow.

Depending on the business logic any permutation of modules under the integrated analytic, prediction, and planning component can be called by *Dispatcher* in a particular data and work flow. Data Persisting and Processing modules stores and process the data upon need. Analysis pavement network, and maintenance prioritization modules are part of strategic, and tactical decision activities, traditionally being part of of PMS and MMS, respectively. And finally there are two adaptive modules of risk assessment framework, and integrated planning module. The former corresponds to climate change adaptation, and the latter refers to adaptiveness in the actual planning activities. This is due to the fact that sources of uncertainty in a plan is not limited to the climate change factors. Any expected or unexpected changes from internal/external variables requires robustness and real-time behaviours, when the discussion is about being adaptive. *Exchanger* module similar to the *Notifier* is a gateway for accessing the database by the third party applications (a reporting application as an example). Regarding software patterns and tactics, *Dispatcher* follows Lambda software architecture by Yamato et al. (2016) with two approaches of cold, and hot paths referring to the batch processing, and streaming (online or real-time) to support variety in work flow scenarios. Regarding input data, data integration component refers to data gathering from various data sources and has four main data sources namely Configuration data including but not limited to road network data, Blue Spot model introduced by Axelsen and Larsen (2014), static and dynamic sources of data like historical or live sensor-based data, respectively.

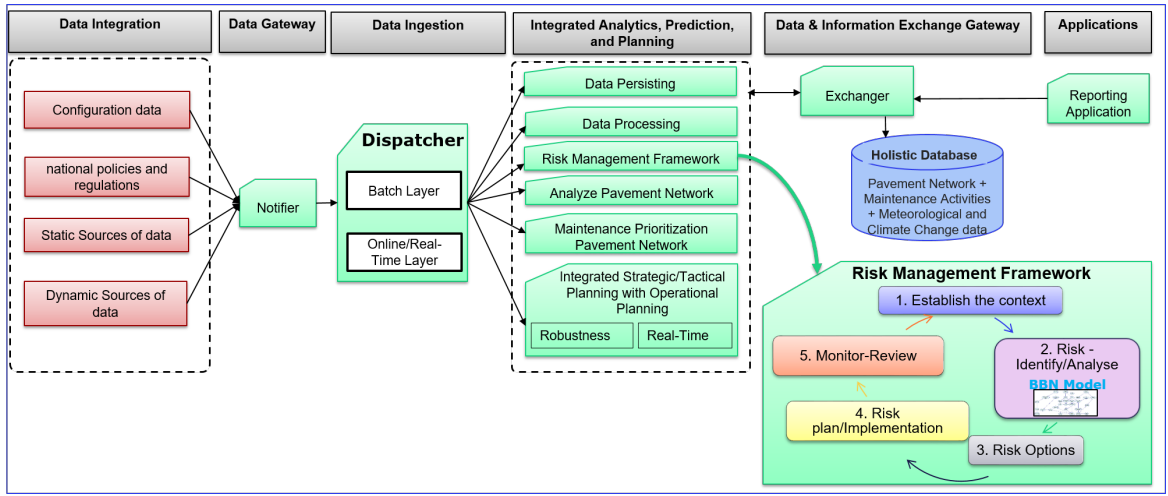

Fig. 1. The Proposed Software Architecture of TDADSS-IPM

As one step towards implementation of such TDADSS-IPM, this paper focuses on modeling of the Risk Assessment Module as part of a risk management framework in the rest of the paper. In the era of climate change, a risk management framework is a stand-alone module or envisioned to be part of a DSS mostly in an iterative manner according to Palutikof et al. (2019). The authors proposed a risk assessment framework with five-steps of *establish the context*, *risk analysis*, *risk options*, *risk implementation*, and *monitor/review*, where our TDADSS-IPM also follows the same framework zoomed in figure 1, and the specific focus in this paper is on the step 2: Risk Analysis/Assessment.

## 3. Risk Management Framework

Regarding *Step 1 - Establish the Context*, the risk assessment on the impact of weather conditions on the Danish road asset is applied in collaboration with the Danish Road Directorate-DRD (Vejdirektoratet). Current practice in DRD is to consider an expectation of medium impact of weather conditions in the next maintenance planning time horizon. However, the importance of knowledge on how weather condition affects the condition of road assets has bee pointed by maintenance decision-maker in the DRD, which such knowledge necessitates a risk assessment. Regarding *Step 2 - Risk Assessment*, Bayesian Belief Network (BBN) proposed by Lam and Bacchus (1994) as a probabilistic graphical model is our candidate approach. Motivation to choose BBN is due to various reasons: i) support the bottom-up approach by Willows et al. (2003) (versus top-down) where asset condition is part of the risk assessment input ii) Ability to quantify the probability of occurrence of asset damages as result of extreme events iii) capability to be upgraded as a dynamic risk assessment module to absorb new events. Prior presenting our conceptual risk assessment model, Table 1 reviews the studies investigated the impact of extreme weather condition on the road transportation. Last row refers to our study in this paper.

### 3.1. Conceptual Risk Assessment model

Figure 2 represents our customized BBN model for the Danish Road network via 28 nodes/variables. Extreme precipitation, extreme temperature, sea level rise, and zero-point crossings all contribute to Denmark's environmental stresses (roots in blue).

Table 1. Investigating the impact of extreme weather conditions on the road transportation. * **Common CIDs** include extreme temperature, extreme precipitation. ** **Common consequences** include Flooding, Landslide, Mudflow, Erosion of construction materials, damage to pavement, damage to bridge and culvert, water on roads, deterioration of road, bridge collapse.

| Study | Climate Impact Drivers | Consequences | Method of study | Risk Assessment |
|---|---|---|---|---|
| TRB, 2008 PETERSON (2008) | Heat wave, Arctic temperature, Sea level rise, Storm and hurricanes | Common consequences**, debries on the road, Thermal tension, pavement integrity | Literature review, Workshop with climate expert | Qualitative |
| IRWIN, 2008 Irwin and Finkel (2008) | Winter Index including: Temperature, Precipitation, Heat wave, zero crossing | Common consequences** | Data collection and data base | Quantitative, IRWIN database, maintenance, cost benefit analysis |
| RIMAROC, 2010 | Common CIDs*, Sea level rise, Heat wave, Snowfall, Frost, Zero-crossing, Wind, Fog | Loss of safety on the road, Costs for the road reconstruction, Financial costs, Loss of confidence, Impact on the environment | Literature review, Workshop with climate expert | Qualitative: 7-step RIMAROCC framework |
| SWAMP, 2010 Larsen et al. (2010) | Precipitation | Flooding | Questionnaire, Data collection (laser scan, rainfall), Chicago design storm (CDS) | Quantitative, Bluespot model |
| EWENT, 2011 Leviäkangas et al. (2012) | Common CIDs*, Wind, Snow, Fog | Common consequences**, Blocking of drainage systems, Vehicle behavior, Falling trees, Slippery road, Accident, Slow traffic, Blocking of road | Media-reported professional literature review | Qualitative: threshold values to critical weather parameters |
| AASHTO, 2012 | Common CIDs*, Sea level, Hurricanes | Common consequences**, dilution of surface salt, wind pressure, Sea level, Encroachment of saltwater, | Literature review | Qualitative |
| NCHRP, Strategic Issues Facing Transportation, 2014 Meyer et al. (2014) | Common CIDs*, Sea level, Hurricanes | Similar to ASHTO 2012 | Literature review | Qualitative |
| ROADAPT, 2015 Bles et al. (2016) | Common CIDs* | Flooding of road surface, Erosion of road embankments and foundations, Landslips and avalanches, Loss of road structure integrity, Loss of pavement integrity, Loss of driving ability due to extreme weather events, the Reduced ability for maintenance | Literature review, Historical data | Qualitative: Quick scan method and steps of sixth and seventh |
| CEDR, 2016 Axelsen et al. (2016) | Common CIDs*, Sea level rise, Heat wave, Snowfall, Frost, Zero-crossing, Wind | Common consequences** | Quick scan, Inventory along chosen road, historical data | Qualitative |
| Kulkarni and Shafaei, 2018 Kulkarni and Shafei (2018) | Common CIDs* | Common consequences** | Literature review, Historical data | Quantitative: Bayesian Belief Network |
| This study, 2022 | Common CIDs*, Sea level rise, Zero-crossing, | Common consequences**, Blocking of drainage systems | Literature review, Historical data | Quantitative: Bayesian Belief Network |

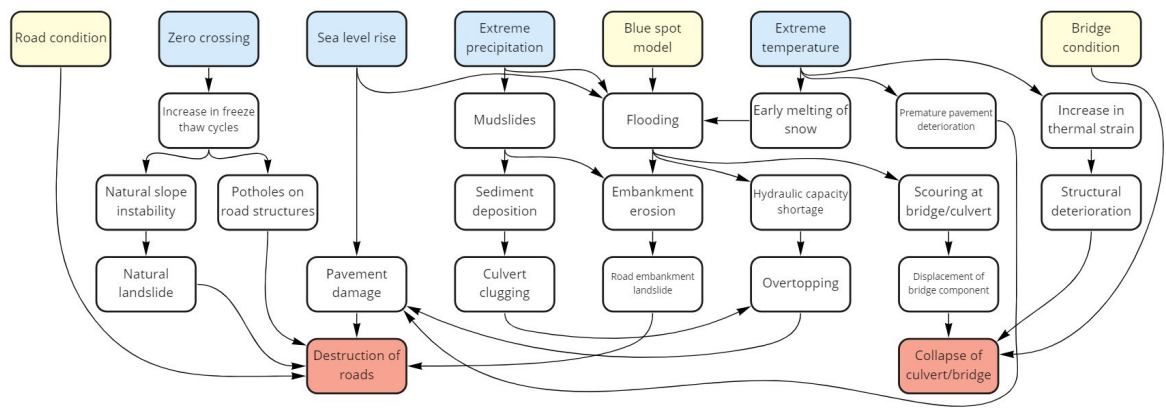

Fig. 2. BBN model proposed by Expert opinion and for the Danish Road Network to study climate change's impact on road transportation assets

Excessive precipitation may result in mudslides and floods. Additionally, extreme temperatures may result in early snow melt, premature pavement damage, and an increase in thermal strain. Extreme precipitation and early snowmelt are two of the primary causes of floods. Sea level rise leads to pavement damage, and flooding. To assess the flood's importance, a Blue Spot Model by Axelsen and Larsen (2014) was employed to pinpoint the critical area. Flooding may have many immediate implications, including scouring of bridges and culverts, erosion of road embankments, and a decrease in the hydraulic capacity of drainage systems. Increased freeze-thaw cycles, as a result of more zero-crossing days, might result in potholes in road structures and natural slope instability. As a consequence, pavements may be destroyed. Additionally, increasing the number of zero crossing days may raise the likelihood of crossing both the dew point and road temperature lines. Similar relations can be derived for bridges collapses. Additionally, the existing state of roads and bridges, as well as the risk level indicated by the blue spot model, have an impact on their ability to endure severe weather events. Separate variables are added into the constructed BBN to accommodate this.

## 3.2. Unconditional and Conditional Probabilities

The probability relationship among parent/s and child/s can be determined either by expert judgment, calculation, direct measurements, stakeholder judgment, or the result of another model as mentioned in Daniel and Iswarani (2020). The unconditional and conditional probabilities in our models are shown in tables 2, 3, and 4. All the roots in the conceptual model are considered as unconditional properties (not having any parent). Unconditional water-related variables are estimated and parameterized by a latest report published by the Danish Meteorological Institute (DMI) which describes the future development of the climate change factors in the country by Pedersen et al.. Accordingly, probability of occurrence of extreme Precipitation and Sea(water) level are calculated by equation 1 proposed by Hershfield (1973). T and r refer to the return period and number of years.

$$U = 1 - (1 - 1/T)^r \qquad (1)$$

Table 2. Unconditional and Conditional probabilities in road deterioration

| Extreme precipitation | | Sea level rise | | Extreme temperature | | Zero crossing | |
|---|---|---|---|---|---|---|---|
| Yes | 0.63 | Yes | 0.63 | Yes | 0.40 | Yes | 0.22 |
| No | 0.37 | No | 0.37 | No | 0.60 | No | 0.78 |

| Mudslides | | | Early melting of snow | | | Sediment deposition | | | Hydraulic capacity shortage | | | Scouring at bridge/culvert | | |
|---|---|---|---|---|---|---|---|---|---|---|---|---|---|---|
| Extreme precipitation | Yes | No | Extreme temperature | Yes | No | Mudslides | Yes | No | Flooding | Yes | No | Flooding | Yes | No |
| Yes | 0.20 | 0.00 | Yes | 0.90 | 0.00 | Yes | 0.70 | 0.10 | Yes | 0.60 | 0.10 | Yes | 0.60 | 0.10 |
| No | 0.80 | 1.00 | No | 0.10 | 1.00 | No | 0.30 | 0.90 | No | 0.40 | 0.90 | No | 0.40 | 0.90 |

| Increase in thermal strain | | | Culvert cluggin | | | Premature pavement deterioration | | | Road embankment landslide | | | Displacement of bridge component | | |
|---|---|---|---|---|---|---|---|---|---|---|---|---|---|---|
| Extreme temperature | Yes | No | Sediment deposition | Yes | No | Extreme temperature | Yes | No | Embankment erosion | Yes | No | Scouring at bridge/culvert | Yes | No |
| Yes | 0.30 | 0.01 | Yes | 0.40 | 0.01 | Yes | 0.30 | 0.20 | Yes | 0.80 | 0.10 | Yes | 0.70 | 0.30 |
| No | 0.70 | 0.99 | No | 0.60 | 0.99 | No | 0.70 | 0.80 | No | 0.20 | 0.90 | No | 0.30 | 0.70 |

| Structural deterioration | | | Increase in freeze and thaw cycles | | | Potholes on road structures | | | Natural slope instability | | | Natural landslide | | |
|---|---|---|---|---|---|---|---|---|---|---|---|---|---|---|
| Increase in thermal strain | Yes | No | Zero crossing | Yes | No | Increase in freeze and thaw cycles | Yes | No | Increase in freeze and thaw cycles | Yes | No | Natural slope instability | Yes | No |
| Yes | 0.40 | 0.30 | Yes | 0.70 | 0.30 | Yes | 0.40 | 0.20 | Yes | 0.40 | 0.30 | Yes | 0.90 | 0.10 |
| No | 0.60 | 0.70 | No | 0.30 | 0.70 | No | 0.60 | 0.80 | No | 0.60 | 0.70 | No | 0.10 | 0.90 |

| Bluespot model | | Road condition | | Bridge condition | | Embankment erosion | | | | Overtopping | | | | |
|---|---|---|---|---|---|---|---|---|---|---|---|---|---|---|
| | | | | | | Mudslides | Yes | | No | | Hydraulic capacity shortage | Yes | | No |
| Low | 0.30 | Very good | 0.20 | Proper | 0.50 | Flooding | Yes | No | Yes | No | Culvert clugging | Yes | No | Yes | No |
| Medium | 0.50 | Good | 0.37 | Need to repair | 0.37 | Yes | 0.70 | 0.60 | 0.30 | 0.01 | Yes | 0.99 | 0.90 | 0.10 | 0.01 |
| High | 0.20 | Towards bad | 0.57 | Must be limited | 0.57 | No | 0.30 | 0.40 | 0.70 | 0.99 | No | 0.01 | 0.10 | 0.90 | 0.99 |
| | | Bad | 0.07 | Not usable | 0.07 | | | | | | | | | | |

| Pavement damage | | | | | | | | Collapse of culvert/bridge | | | | | | | | | | | | | |
|---|---|---|---|---|---|---|---|---|---|---|---|---|---|---|---|---|---|---|---|---|---|
| Overtopping | Yes | | | | No | | | Displacement of bridge component | Yes | | | | | No | | | | | | | |
| Premature pavement deterioration | Yes | | No | | Yes | | No | Structural deterioration | Yes | | | | No | Yes | | | | No | | | |
| Sea level rising | Yes | No | Yes | No | Yes | No | Yes | No | Bridge condition | G1 | G2 | G3 | G4 | G5 | G1 | G2 | G3 | G4 | G5 | G1 | G2 | G3 | G4 | G5 | G1 | G2 | G3 | G4 | G5 |
| Yes | 0.7 | 0.7 | 0.3 | 0.3 | 0.2 | 0.2 | 0.01 | 0.01 | Yes | 0.2 | 0.4 | 0.6 | 0.8 | 0.99 | 0.1 | 0.2 | 0.4 | 0.6 | 0.9 | 0.01 | 0.1 | 0.3 | 0.4 | 0.6 | 0.01 | 0.01 | 0.1 | 0.2 | 0.3 |
| No | 0.3 | 0.3 | 0.7 | 0.7 | 0.8 | 0.8 | 0.99 | 0.99 | No | 0.8 | 0.6 | 0.4 | 0.2 | 0.01 | 0.9 | 0.8 | 0.6 | 0.4 | 0.1 | 0.99 | 0.9 | 0.7 | 0.6 | 0.4 | 0.99 | 0.99 | 0.8 | 0.8 | 0.7 |

Number of years is set to 100, as the average life expectancy of a road asset is usually above 50 years. Regarding return period, DMI considers the dependability of climate change to the greenhouse gases(scenarios of RCP4.5 and RCP8.5), where the RCP8.5 is relevant when the number of years is considered as 100. Return period is related to the accepted risk for road assets, usually set as 100 years according to the standards of road construction, too. Accordingly, Extreme Precipitation and Sea(water) Level rise are calculated by above formula as 0.63 and 0.37 in case of yes, and no, respectively. Extreme Temperature and Zero crossing are taken from a similar study on future of climate changes predicted by Swedish Meteorological and Hydrological Institute (SMHI), and are set to 0.40 and 0.60, and 0.22 and 0.78 in case of yes, and no, in the order given. Road condition classification (very good, good, towards bad, bad) is provided by stakeholder (the DRD), and Bridge condition classification (proper, need to repair, must be limited, not usable) is according to the Bridge Danish Management system (Danbro), both elicited from visional inspection. Blue Spot levels(low, medium, high) comes from the original model, and values are assumed and given by the expert opinion as 0.30, 0.50, and 0.20 for potential flooding-recognized points, respectively. All the other values for the conditional probabilities (which have at least one parent), are assumed and given by the hydraulic domain expert as initial belief for the made BBN.

Table 3. Conditional Probabilities for deterioration of Roads

| Pavement damage | Yes | | | | | | | | | | | | | | | | | | | | | | | |
|---|---|---|---|---|---|---|---|---|---|---|---|---|---|---|---|---|---|---|---|---|---|---|---|---|
| Road embankment landslide | Yes | | | | | | | | | | | No | | | | | | | | | | | | |
| Natural landslide | Yes | | | | | | No | | | | | | Yes | | | | | | No | | | | | |
| Pothole on road structures | Yes | | | No | | | Yes | | | No | | | Yes | | | No | | | Yes | | | No | | |
| Road Condition | good | fair | poor | good | fair | poor | good | fair | poor | good | fair | poor | good | fair | poor | good | fair | poor | good | fair | poor | good | fair | poor |
| YES | 0.99 | 0.99 | 0.99 | 0.90 | 0.95 | 0.99 | 0.90 | 0.95 | 0.99 | 0.90 | 0.95 | 0.99 | 0.70 | 0.80 | 0.99 | 0.60 | 0.70 | 0.90 | 0.40 | 0.60 | 0.90 | 0.30 | 0.50 | 0.80 |
| NO | 0.01 | 0.01 | 0.01 | 0.10 | 0.05 | 0.01 | 0.10 | 0.05 | 0.01 | 0.10 | 0.05 | 0.01 | 0.30 | 0.20 | 0.01 | 0.40 | 0.30 | 0.10 | 0.60 | 0.40 | 0.10 | 0.70 | 0.50 | 0.20 |

| Pavement damage | No | | | | | | | | | | | | | | | | | | | | | | | |
|---|---|---|---|---|---|---|---|---|---|---|---|---|---|---|---|---|---|---|---|---|---|---|---|---|
| Road embankment landslide | Yes | | | | | | | | | | | No | | | | | | | | | | | | |
| Natural landslide | Yes | | | | | | No | | | | | | Yes | | | | | | No | | | | | |
| Pothole on road structures | Yes | | | No | | | Yes | | | No | | | Yes | | | No | | | Yes | | | No | | |
| Road Condition | good | fair | poor | good | fair | poor | good | fair | poor | good | fair | poor | good | fair | poor | good | fair | poor | good | fair | poor | good | fair | poor |
| YES | 0.95 | 0.99 | 0.99 | 0.90 | 0.99 | 0.99 | 0.95 | 0.99 | 0.99 | 0.90 | 0.99 | 0.99 | 0.70 | 0.80 | 0.90 | 0.60 | 0.70 | 0.80 | 0.20 | 0.50 | 0.60 | 0.01 | 0.10 | 0.50 |
| NO | 0.05 | 0.01 | 0.01 | 0.10 | 0.01 | 0.01 | 0.05 | 0.01 | 0.01 | 0.10 | 0.01 | 0.01 | 0.30 | 0.20 | 0.10 | 0.40 | 0.30 | 0.20 | 0.80 | 0.50 | 0.40 | 0.99 | 0.90 | 0.50 |

Table 4. Conditional Probabilities Assumed for "Collapse of Bridges", Bridge Condition Grades: G1: Proper condition, G2: Bridge needs to be repaired, G3: Bridge needs to be repaired basically, G4: Traffic must be limited, G5: Bridge is not useable

| Displacement of bridge component | Yes | | | | | | | | | | No | | | | | | | | | |
|---|---|---|---|---|---|---|---|---|---|---|---|---|---|---|---|---|---|---|---|---|
| Structural deterioration | Yes | | | | | No | | | | | Yes | | | | | No | | | | |
| Bridge condition | G1 | G2 | G3 | G4 | G5 | G1 | G2 | G3 | G4 | G5 | G1 | G2 | G3 | G4 | G5 | G1 | G2 | G3 | G4 | G5 |
| Yes | 0.0 | 0.0 | 0.0 | 0.0 | 0.9 | 0.0 | 0.0 | 0.0 | 0.0 | 0.0 | 0.0 | 0.0 | 0.0 | 0.0 | 0.0 | 0.0 | 0.0 | 0.0 | 0.0 | 0.0 |
|  | 0.2 | 0.4 | 0.6 | 0.8 | 0.9 | 0.1 | 0.2 | 0.4 | 0.6 | 0.9 | 0.1 | 0.1 | 0.3 | 0.4 | 0.6 | 0.1 | 0.1 | 0.1 | 0.2 | 0.3 |
| No | 0.0 | 0.0 | 0.0 | 0.0 | 0.0 | 0.0 | 0.0 | 0.0 | 0.0 | 0.0 | 0.9 | 0.0 | 0.0 | 0.0 | 0.0 | 0.9 | 0.9 | 0.0 | 0.0 | 0.0 |
|  | 0.8 | 0.6 | 0.4 | 0.2 | 0.1 | 0.9 | 0.8 | 0.6 | 0.4 | 0.1 | 0.9 | 0.9 | 0.7 | 0.6 | 0.4 | 0.9 | 0.9 | 0.9 | 0.8 | 0.7 |

## 4. Results

Ultimately, after collecting data through calculation, literature review, and expert judgment to fill the probabilities table (2, 3, 4), the probability of road deterioration and bridge collapse is calculated via the BBN formula represented in equation 2, where PP are parents of $X_i$ node.

$$p(X1, X2, X3, \ldots, Xn) = \prod_{i=0}^{n} p(Xi | PP(Xi)) \qquad (2)$$

Our conceptual risk model is modeled via GeNIe Modeler introduced in BayesFusion (2017) which is freely available for academic purposes is employed. The result of run model on the initial assumed probabilities is shown in figure 3. As can be seen in this figure, the probability of road deterioration and bridge collapse, respectively, is 34% and 48% based on the current condition of roads and bridges, the flooding zone determined by the blue spot model, and the probability of the occurrence of extreme events such as heatwaves, precipitation, sea level rises, and zero crossings. At a glance, it can be recognized from the model that the condition of bridges is more endanger of collapse. As discussed,the collapse of bridges is influenced directly by structural deterioration and displacement of the bridge component, which is affected by increasing thermal heat and scouring at one level above. In order to be adapting to these issues, some actions in different levels of emergency can be undertaken. For instance, as the urgent action, the most prioritized structures for remediations should be determined by monitoring the bridges and culverts. Consequently, expanding the capacity of bridges and culverts and increasing the expansion joints can be considered. It is also more reliable to revise the regulations, and design criteria in guidelines should be revised. Making such BBN with the aforementioned assumed probabilities, provide an opportunity to test various scenarios. Here to demonstrate our purposed BBN, first, we represent the result of the initial assumptions in figure 3. Then to evaluate the impacts of roads condition, bridge conditions, and the blue spot model with the bottom-up method, the unconditional variables (extreme precipitation, sea-level rise, extreme precipitation, and zero-crossing) are considered as predefined events, and all are set to 1 (the worst-case scenario). Figure 5 and 6 demonstrate probability of road destruction and collapse of bridges under worst case scenario over different states of blue spot model, and different categories of road and bridge conditions. Figure 4 is representative of worst case scenario, when road and bridge conditions are in categories of "Bad" and "Not usable". In this specific scenario, the most probable condition that leads to road deterioration is 73%, and the corresponding effects of the bridge conditions are remarkable with the amount of 86% that enlighten the importance of bridge main- tenance. This is apparent as can be seen under any condition of flooding resulted by bluespot model, the difference between the risk of bridge collapse with not useable and proper condition is approximately around 44% (averaged by 81% - 35%, 83% - 38%, and 86% - 44%, in case of low,medium, and high bluespot, respectively).

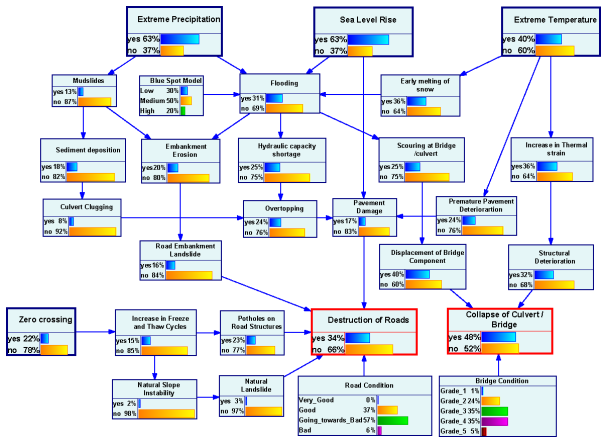
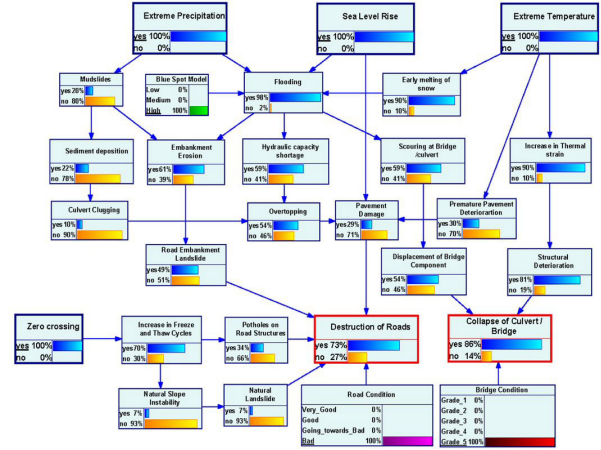

Fig. 3. Results of Assumed BBN model of Danish roads against the climate change phenomena

Fig. 4. Results of BBN model of Danish roads in Worst Case Scenario of the climate change factors

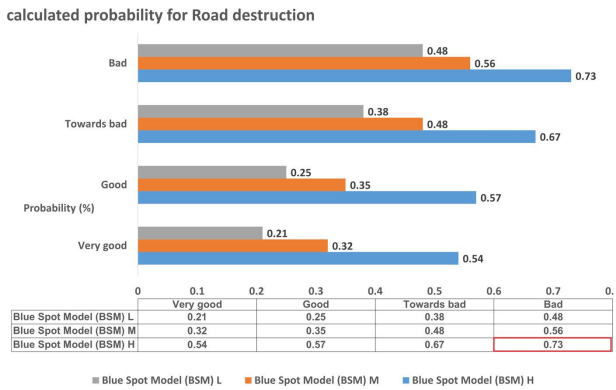
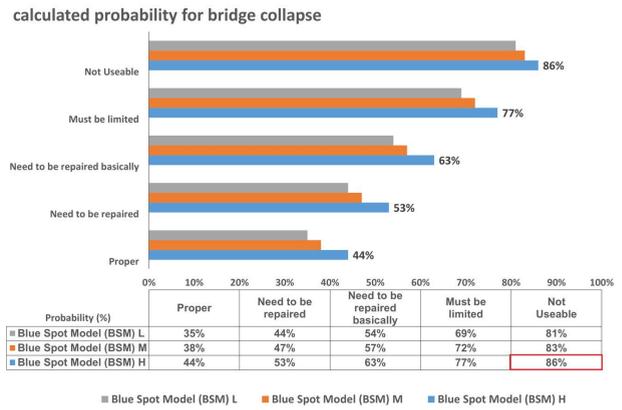

Fig. 5. Probabilities of the road deterioration under happening extremes and different states of blue spot, road, and bridge conditions

Fig. 6. Probabilities of the bridge collapse under happening extremes and different states of blue spot, road, and bridge conditions

## 5. Conclusion

Some immediate pay-offs follow with a holistic TDADSS-IPM, remedying miscommunication and unifying strategy and management from the various departments, and all levels of strategic, tactical, and operational decisions. Moreover, the solution will present itself with the possibility of building features that is reliant on events from both PMSs and MMSs. This is achievable via ability to react to the uncertainties via risk assessment framework, and robust and real-time planning. Moreover, efficient planning and fast reaction means less road maintenance uptime, focus on environmental impact in addition to effectivizing road maintenance planning, means a stronger road network for the public. Moreover, the modeled BBN approach facilitates the transformation of prescriptive knowledge into descriptive knowledge in the era of prediction of extreme events. In light of this, it is critical to recognize that the prioritized risks and the hierarchical relationships between influential factors and the vulnerable assets of roads against climate change extend beyond precise results. The BBN model's flexible structure enables policymakers to revise their beliefs in light of real-time events and train the model according to newly updated information as it becomes available. Gathering the views of road experts and policymakers can produce a valuable source of information based on common sense. It can be accomplished in the first place by preparing a questionnaire and distributing it to the experts. Moreover, making the BBN model as a dynamic model is an immediate interest.